\documentclass[journal]{IEEEtran}
\usepackage{booktabs, makecell, multirow, tabularx}
\usepackage{cite}
\usepackage{amsmath,amssymb,amsfonts}
\usepackage{algorithmic}
\usepackage{graphicx}
\usepackage{textcomp}
\usepackage{hyperref}
\usepackage[utf8]{inputenc}
\usepackage[T1]{fontenc}
\usepackage{times}
\usepackage{epsfig}
\usepackage{multicol,multirow}
\usepackage{booktabs}
\usepackage{bigstrut}
\usepackage{floatrow}
\usepackage{rotating}
\usepackage{adjustbox}
\usepackage{mathtools}
\usepackage{longtable}
\usepackage{array}

\begin{document}

\title{Focal Modulation and Bidirectional Feature Fusion Network for Medical Image Segmentation}

\author{Moin Safdar, Shahzaib~Iqbal, \IEEEmembership{Member, IEEE}, Imran Razzak, \IEEEmembership{Senior Member, IEEE} and Tariq~M.~Khan, \IEEEmembership{Member, IEEE}, Thantrira Porntaveetus and Hamid Alinejad-Rokny, \IEEEmembership{Senior Member, IEEE}
\thanks{Shahzaib, is with the Department of Computing, Abasyn University Islamabad Campus (AUIC), Islamabad, Pakistan (e-mail: shahzaib.iqbal91@gmail.com).}
\thanks{Mubeen Ghafoor is with School of Computer Science and Informatics, De Montfort University, UK (e-mail:mubeen.ghafoor@dmu.ac.uk). }
\thanks{Tariq M. Khan is with Naif Arab University for Security Sciences, Riyadh, KSA (e-mail: tariq045@gmail.com).}
\thanks{Imran Razzak is with the School of Computer Science and Engineering, University of New South Wales, Sydney, NSW, Australia (e-mail: imran.razzak@unsw.edu.au).}
\thanks{ Thantrira Porntaveetus is with the Center of Excellence in Precision Medicine and Digital Health, Chulalongkorn University, Thailand.}
\thanks{Hamid Alinejad-Rokny is with the School of Biomedical Engineering, UNSW Sydney, Australia; and Visiting Scholar (Collaborative Projects), Center of Excellence in Precision Medicine and Digital Health, Chulalongkorn University, Thailand; (e-mail: h.alinejad@unsw.edu.au)} 
}%

\maketitle
\begin{abstract}
Medical image segmentation is essential for clinical applications such as disease diagnosis, treatment planning, and disease development monitoring because it provides precise morphological and spatial information on anatomical structures that directly influence treatment decisions. Convolutional neural networks significantly impact image segmentation; however, since convolution operations are local, capturing global contextual information and long-range dependencies is still challenging. Their capacity to precisely segment structures with complicated borders and a variety of sizes is impacted by this restriction. Since transformers use self-attention methods to capture global context and long-range dependencies efficiently, integrating transformer-based architecture with CNNs is a feasible approach to overcoming these challenges. To address these challenges, we propose the Focal Modulation and Bidirectional Feature Fusion Network for Medical Image Segmentation, referred to as FM-BFF-Net in the remainder of this paper. The network combines convolutional and transformer components, employs a focal modulation attention mechanism to refine context awareness, and introduces a bidirectional feature fusion module that enables efficient interaction between encoder and decoder representations across scales. Through this design, FM-BFF-Net enhances boundary precision and robustness to variations in lesion size, shape, and contrast. Extensive experiments on eight publicly available datasets, including polyp detection, skin lesion segmentation, and ultrasound imaging, show that FM-BFF-Net consistently surpasses recent state-of-the-art methods in Jaccard index and Dice coefficient, confirming its effectiveness and adaptability for diverse medical imaging scenarios.
\end{abstract}


\section{Introduction}
\IEEEPARstart{M}{edical} image segmentation plays a crucial role in the recognition and differentiation of anatomical features, lesions, and diseases within various types of medical images. Numerous clinical applications, such as disease diagnosis, therapy planning, and disease progression monitoring, are heavily dependent on this process \cite{ soomro2016automatic,naqvi2019automatic,khan2020shallow,khan2020semantically}. Performing accurate segmentation of medical images allows medical professionals to obtain essential information about the morphological and spatial properties of tumors, organs, and other areas of interest. Medical image segmentation divides the image into several significant sections according to textures, pixel intensities, and other significant characteristics \cite{abdullah2021review,khan2021residual,khan2021rc,iqbal2022g}. This makes it possible to evaluate diseases separately from their surroundings. In recent years, convolutional neural networks (CNNs) have made significant improvements in medical image segmentation \cite{arsalan2022prompt,iqbal2022recent, qayyum2023semi,khan2023simple,khan2023retinal,iqbal2023robust}. Due to their ability to capture and interpret local spatial information, CNNs have become the preferred method for medical image segmentation \cite{iqbal2023mlr,iqbal2023fusion,khan2023feature,naveed2024pca,iqbal2023ldmres}. \\

CNN-based segmentation techniques are adequate; however, they have some inherent limitations. CNNs have difficulties with size and borders, which affect a certain level of segmentation accuracy \cite{mazher2024self,naveed2024ra,javed2024advancing,matloob2024lmbis,khan2024lmbf,javed2024region,iqbal2024tesl,iqbal2024euis}.\\ 

 \begin{itemize}
 \item Differences in image quality, contrast, and anatomical structures between various imaging modalities, such as MRI, CT, ultrasound, and X-rays, make medical image segmentation even more difficult \cite{iqbal2025tbconvl,naveed2024ad,farooq2024lssf}.
\item Convolutional operations are localized and usually use fixed-sized filters to extract features from small regions of the image. This makes capturing long-range dependencies and global context challenging and essential to accurately segment medical images, mainly when working with structures that vary in size, shape, and texture \cite{khan2024esdmr,mehmood2024lvs,xu2025edge,naveed2025fm}. 
\item Regarding medical imaging, where abnormalities and lesions may take different forms in different patients and imaging modalities, segmentation performance may be hampered by the inability to capture such global information \cite{khan2025role,mehmood2025lfra}. 
\item Border difficulties occur when CNNs incorrectly segment objects with complex or irregular shapes because they often fail to capture fine features within the boundaries of anatomical structures \cite{xu2025entropy,khan2025novel}. 
\item Scaling challenges present additional difficulty because anatomical structures can differ significantly in size. CNNs trained on a specific scale might not generalize well to structures of different sizes, which could lead to inaccurate segmentation \cite{tajbakhsh2020embracing}. 
 \end{itemize}

 This is crucial in a clinical environment where precise segmentation might directly affect treatment choices, such as in cardiology for blood artery segmentation or oncology for tumor borders \cite{ doolub2023artificial}. 
Thus, there is an increasing demand for approaches that accurately represent local and global interactions in medical images. To address these issues, CNNs must be able to capture contextual details at a local and global level, which is crucial to segment structures of different sizes and shapes \cite{hafiz2020survey}. The small receptive field of traditional CNNs makes them excellent at capturing local features, but they sometimes struggle to capture long-range dependencies \cite{liu2023improving}. The model can be improved to recognize structures on various scales by adding multiscale feature extraction layers \cite{zhang2024multi}. Additionally, methods such as boundary refinement modules or post-processing can increase border accuracy by fine-tuning the borders of segmented regions \cite{jia2025fbsm}.\\

Integrating transformer-based architectures with CNNs is one potential strategy to overcome these challenges. In order to better reflect global context and long-range interdependence, attention methods enable models to concentrate on critical regions. Transformers can be used with CNNs to capture local and global features, eliminating border and scale difficulties \cite{mehmood2024retinalitenet}. Transformers are well-known for their ability to estimate connections across entire images effectively.\\


Developing segmentation models that can achieve high accuracy while retaining the simplicity of computing is becoming increasingly important as technology progresses. This is important for applications operating in resources-limited locations where data availability and processing power may be restricted, such as mobile devices or remote healthcare settings \cite{manda2010implementing}. New methods that can balance accuracy, computing efficiency, and generalizability across many imaging modalities will become more crucial as medical image segmentation develops and advances healthcare \cite{ rayed2024deep}.
In conclusion, the topic of medical image segmentation is rapidly developing and has significant implications for healthcare. More precise treatment strategies, more accurate disease diagnoses and better patient outcomes are possible with accurate segmentation. Segmentation models will probably become even more crucial to clinical processes as technology advances, improving the capacity of healthcare providers to provide excellent treatment in various medical situations.\\

The following are the main contributions of this paper.\\

\begin{itemize}
  
\item Combining CNN- and transformer-based components to take advantage of both local feature extraction and global context awareness, achieving high segmentation accuracy.

\item Introducing a focal modulation-based convformer attention block (FMCAB) to modulate feature flow between the encoder and decoder, enhancing local-global context integration, and improving segmentation accuracy.

\item Developing a bidirectional feature fusion module (BiFFM) that combines feature information from each encoder stage with the decoder, utilizing skip connections to enrich context at every stage for better segmentation performance.

\item Using EfficientNetV2S1 as the backbone to maximize computational efficiency and aggregated skip connections to aggregate the contextual information extracted at each stage of the encoder-decoder.
  
\end{itemize}

The remaining manuscript is arranged as follows. Recent related research on lightweight medical image segmentation models and retinal feature segmentation techniques is presented in Section \ref{sec: Related Work}. Section \ref{method} contains the specifics of the suggested model, M-BFF-Net. Detailed experiments and M-BFF-Net results are published in Section \ref{experimentalResults}, along with an explanation of the experimental environment. The key findings and conclusions of the suggested study are finally summarized in Section \ref{sec:Conclusions}.

\begin{figure*}[!t]
    \centering
    \includegraphics[width=\textwidth]{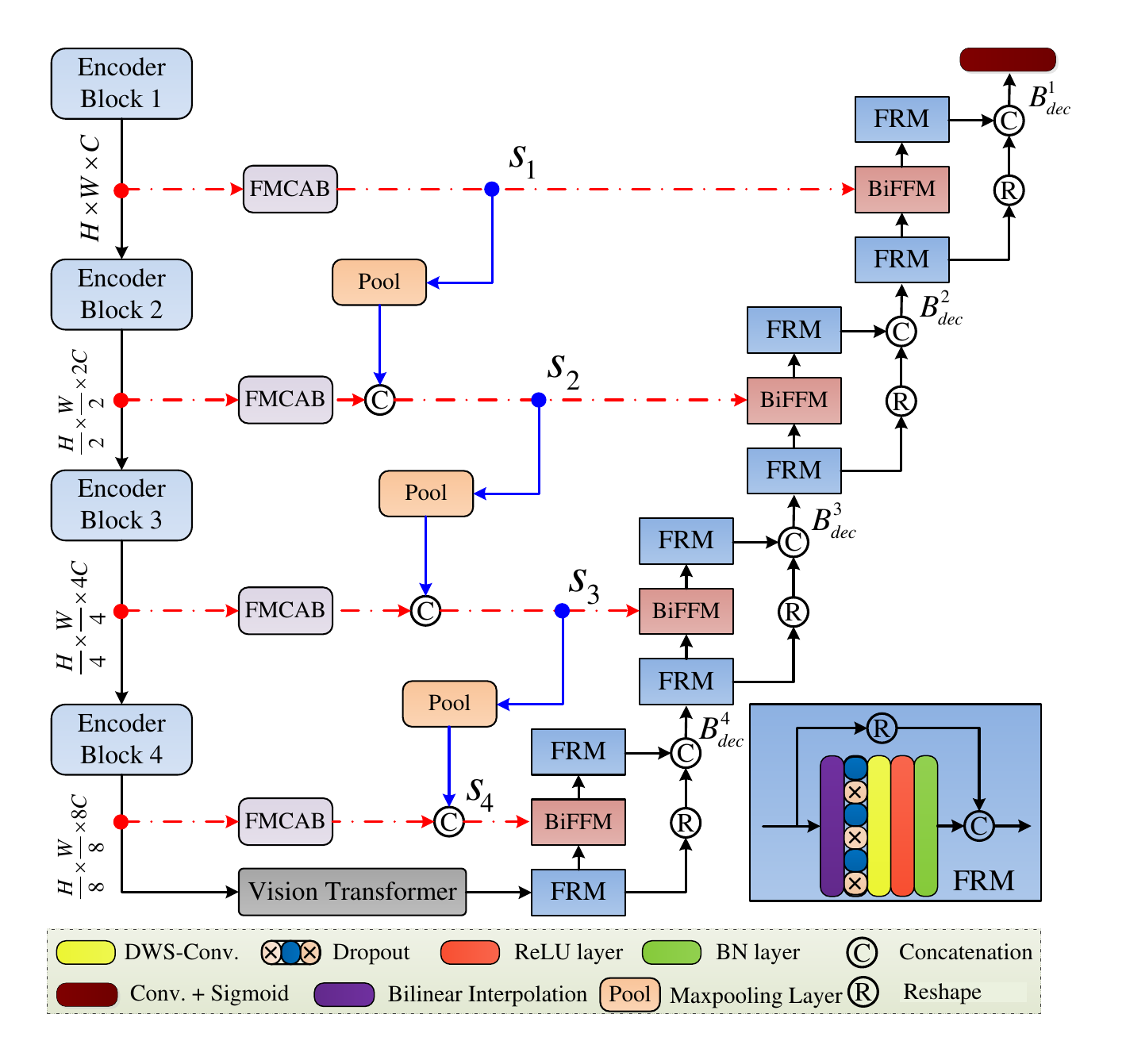}
    \caption{Overview of the proposed M-BFF-Net architecture for medical image segmentation. The model integrates convolutional and transformer-based modules to capture both local and global features. It consists of four encoder blocks, each followed by a Focal Modulation-based ConvFormer Attention Block (FMCAB). The encoded features are refined through a Vision Transformer and progressively decoded using Feature Refinement Modules (FRMs) and Bidirectional Feature Fusion Modules (BiFFMs). Skip connections and multi-scale fusion enhance the contextual representation and boundary precision of segmentation outputs.}
    \label{fig:model}
\end{figure*}

\section{Related Work}
\label{sec: Related Work}

CNNs have seen widespread application in the field of MIS due to their powerful feature representation capability. UNet \cite{ronneberger2015u} is a seminal architecture in this domain and has achieved competitive results on various MIS tasks. Several variations of UNet have been introduced, including Dense-UNet \cite{li2018h}, UNet++ \cite{zhou2019unet++}, UNet3+ \cite{huang2020unet}, nnUNet \cite{isensee2021nnu}, and Attention UNet \cite{oktay2018attention}. Certain approaches are customized for particular objectives, such as segmentation of the optic cup and optic disc from fundus images \cite{fu2018joint} or segmentation of COVID-19 lung infection \cite{fan2020inf}. In contrast, transformer-based methods are known for their notable performance for vision tasks \cite{zheng2021rethinking}, \cite{zhang2021rest, wang2108crossformer}, \cite{liu2021swin, strudel2021segmenter, he2021transreid}. ViT \cite{vaswani2017attention} is the pioneering work in introducing transformers for image classification, and DeiT \cite{touvron2021training} proposed several efficient training strategies for effective training of ViT \cite{vaswani2017attention}. Liu et al. introduced the Swin transformer \cite{liu2021swin} as a technique to perform self-attention using local windows for computer vision applications. This approach reduces computational costs while still producing satisfactory results. Some approaches introduced CNNs' design principles \cite{dai2021coatnet, srinivas2021bottleneck}, into transformers to obtain notable performance and resource efficiency. Several works have been proposed that generalize to 2D and 3D MIS tasks \cite{he2023h2former, zhang2021transfuse, xie2021cotr, yan2022after}, such as nnFormer \cite{zhou2021nnformer} and TransUNet \cite{chen2021transunet}. Chen et al. proposed TransUNet \cite{chen2021transunet} as the initial model to use a hybrid CNN-transformer architecture for MIS. This technique combines local CNN features with global contextual transformer features. Swin-UNet \cite{cao2023swin} was developed based on the principles of Swin transformer \cite{liu2021swin}. However, it does not take into account local spatial information, which is essential in segmentation. In response to the transformers' requirement for large amounts of data, UTNet \cite{gao2021utnet} was proposed, integrating self-attention into a CNN to improve MIS.\\

In MIS, CNNs—a type of deep learning model—have been widely used. This is because of their exceptional ability to extract image features effectively. U-Net \cite{ronneberger2015u} is one of the notable architectures that has become a cutting-edge model that demonstrates competitive performance in various MIS tasks. UNet++ \cite{zhou2019unet++}, nnUNet \cite{isensee2021nnu}, UNet3+ \cite{huang2020unet}, Dense-UNet \cite{li2018h}, and Attention U-Net \cite{oktay2018attention} are some of the variations based on U-Net that have been proposed. These variations of the U-Net and customized methods show how flexible and successful CNNs are at handling a wide range of MIS task difficulties, including particular anatomical features, diseases, or imaging modalities.\\

Furthermore, transformer-based methods have demonstrated outstanding performance \cite{zheng2021rethinking, zhang2021rest, liu2021swin}. The use of transformers in image classification was transformed by the ViT architecture \cite{vaswani2017attention}, which demonstrates how effectively they capture global contextual information. ViT performance has been improved by effective training techniques offered by later developments, such as the DeiT model \cite{touvron2021training}. Using self-attention with local windows, the Swin Transformer \cite{liu2021swin} is a remarkable advancement that enables more computationally efficient processing while still producing adequate results. Some methods have combined the architecture of transformers and CNNs to combine their respective advantages. CoatNet \cite{dai2021coatnet} and bottleneck transformers \cite{srinivas2021bottleneck}, for example, improved performance and resource efficiency by incorporating CNN-inspired architectural components into transformers. The potential advantages of transformer-based approaches for MIS can be investigated due to these developments and their ability to perform vision tasks.\\

Numerous methods have been developed in MIS to handle both 2D and 3D problems. These methods comprise a range of approaches and strategies to address issues particular to medical images. These include techniques such as TransUNet \cite{chen2021transunet}, nnFormer \cite{zhou2021nnformer}, and others \cite{he2023h2former, yan2022after}. For MIS, TransUNet \cite{chen2021transunet} was the first to combine the advantages of transformer and CNN architectures. This novel approach takes advantage of transformers' capacity to recognize global contextual features while utilizing CNN's ability to extract local features. UTNet was introduced to alleviate the data-intensive dependence related to transformers \cite{gao2021utnet}. This approach improves performance on MIS tasks by integrating a self-attention mechanism into a CNN architecture. However, due to their complex designs, redundant feature learning, and higher training computing requirements, TransUNet and UTNet are more likely to overfit. The Swin-UNet model \cite{cao2023swin} was presented based on the concepts of the Swin Transformer \cite{liu2021swin}. Local spatial information is crucial to the segmentation process, but is not considered enough.

\begin{figure*}[h]
    \centering
    \includegraphics[width=1\textwidth]{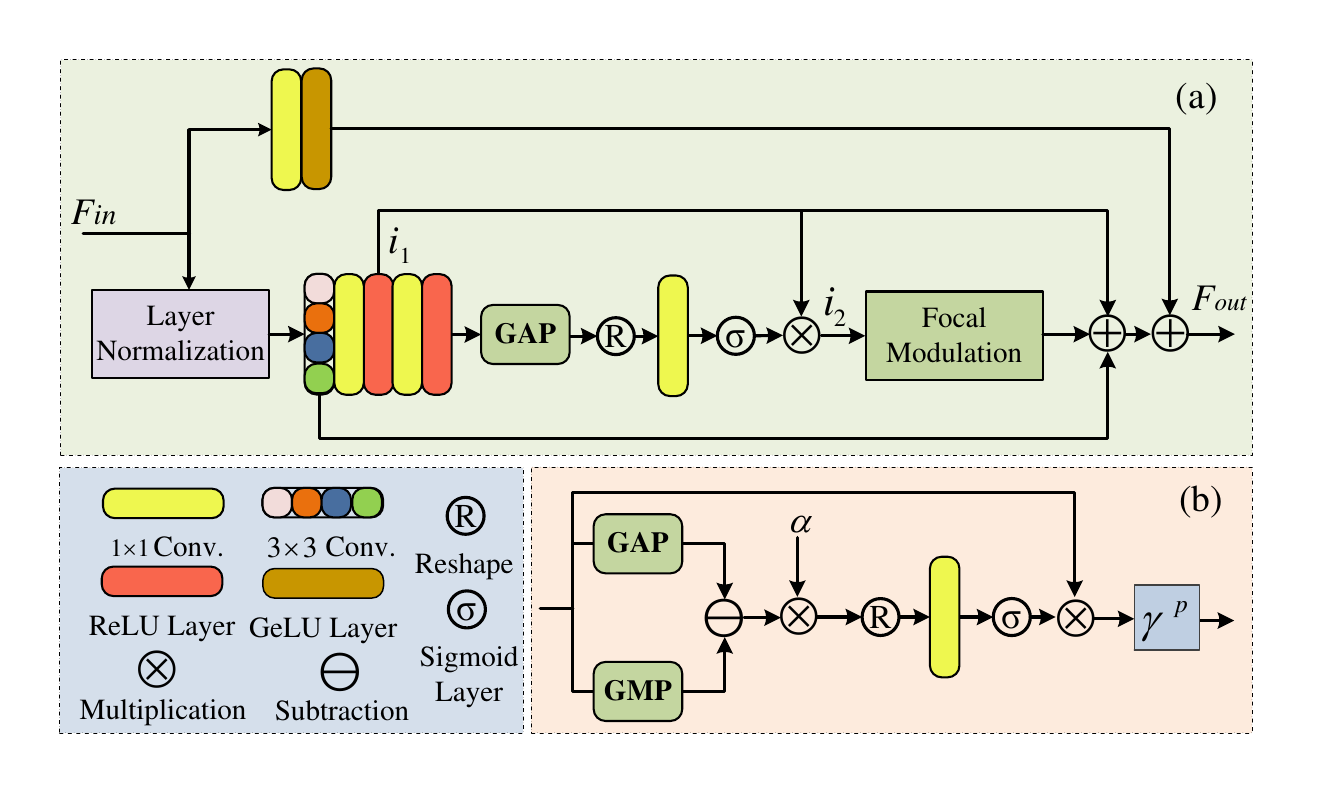}
    \caption{ (a) Architecture of the proposed Focal Modulation-based ConvFormer Attention Block (FMCAB), which combines convolutional and attention mechanisms to enhance spatial and contextual feature representation. The block leverages global average pooling (GAP), reshaping operations, and focal modulation to adaptively refine features. (b) Internal structure of the Focal Modulation (FM) unit, which computes attention weights by fusing outputs from global average pooling (GAP) and global max pooling (GMP), followed by nonlinear transformations for precise focus modulation.}
    \label{fig:model}
\end{figure*}
\section{Proposed Method}\label{method}
The proposed M-BFF-Net, illustrated in Fig.~\ref{fig:model}, consists of five components: the encoder, which is responsible for feature information extraction; the focal modulation-based convformer attention block (FMCAB), which regulates information flow between the encoder and decoder; FMCAB applied to the skip connections from the output of each encoder block, followed by a novel bidirectional feature fusion module (BiFFM) to merge feature information at each decoder stage; the Vision Transformers (ViTs), which leverages the inherent capabilities of transformers to capture extensive dependencies, facilitate flexible interactions between features, and enhance contextual understanding at the bottleneck layer; and the decoder module to reconstruct the feature information. In this section, we comprehensively describe each component. For the encoder of our model, we use EfficientNetV2 \cite{Tan2021}, selected for its superior performance on ImageNet. To manage computational costs, we specifically choose EfficientNetV2S1 as the backbone model. We establish aggregated skip connections within each stage of the encoder and its corresponding decoder block, ensuring that each pair maintains the same feature map dimensions.
In the encoder stage, we have employed four EfficientNetV2S1 encoder blocks denoted $[B_{enc}^{1}, B_{enc}^{2}, B_{enc}^{3}, B_{enc}^{4}]$. Let $f_{in} \in \mathbb{R}$ be the RGB input to the proposed network. The output of the $1^{st}$ skip connection is computed by employing a focal modulation-based convformer attention block $(FMCAB)$ on the first encoder block, as shown in (Eq. \ref{eq:1}).
\begin{equation}
        s_{1} = \texttt{FMCAB}\left ( B_{enc}^{1}(f_{in}) \right )
    \label{eq:1}
\end{equation}

The output of the $n^{th}$ skip connection is computed by employing a focal modulation-based convformer attention block $(FMCAB)$ on the $n^{th}$ encoder block and concatenating it with the skip connection of the previous block ($s_{n-1}$) as shown in (Eq. \ref{eq:2}). A max-pooling operation ($Pool$) is applied on ($s_{n-1}$) to reduce the spatial dimensions of the features. 

\begin{equation}
s_{n} = \texttt{FMCAB}\left ( B_{enc}^{n}(B_{enc}^{(n-1)}) \right ) \copyright \left ( \texttt{Pool}(s_{n-1}) \right )
    \label{eq:2}
\end{equation}
 where $n=2,3, 4$. $\copyright$ is the concatenation. After the feature details are extracted at the encoder stage, a vision transformer-based self-aware attention mechanism is applied to further enhance the feature information and capture long-range dependencies at multiple scales. The final extracted feature information ($f^{enc}$) is computed as described in (Eq.~\ref{eq:3}).

\begin{equation}
        f^{enc}=\texttt{ViT}(B_{enc}^{4})
    \label{eq:3}
\end{equation}

At the decoder stage, the extracted feature information is initially input into the feature reconstruction module (FRM) as described in (Eq. \ref{eq:FRM}), followed by the bidirectional feature fusion module (BiFFM), which also takes $s_{4}$ as a second input to fuse the feature information. 

\begin{equation}
        \texttt{FRM}=[\beta _{n}\left ( \Re\left ( f_{\texttt{DWS}}^{3\times 3}\left ( D_{r}^{0.5}\left ( U_{p}\left ( in \right ) \right ) \right ) \right ) \right )] \copyright [\texttt{Re}\left ( in \right )]
    \label{eq:FRM}
\end{equation}
where $\beta _{n}$ is the batch normalization operation, $f_{\texttt{DWS}}^{3\times 3}$ is depth-wise separable convolution operation with a kernel size ($3\times 3$), $\Re$ is the ReLU activation function $D_{r}^{0.5}$ is the dropout with 0.5 probability, and  $U_{p}$ is the upsamling with bilinear interpolation. Subsequently, a second FRM is applied to this fused information. The output from this second FRM is then concatenated with the output of the first FRM after reshaping to match the spatial context. The primary difference between the two FRM modules is that upsampling is not performed in the second FRM module. The output of the $1^{st}$ decoder block $B_{dec}^{1}$ is computed as described in (Eq.~\ref{eq:4}).
\begin{equation}
        B_{dec}^{1}=\texttt{FRM}\left [ \texttt{BiFFM}  \left \{ \texttt{FRM}(f^{enc}),s_{4} \right \}\right ]\copyright \left [ \texttt{Re}\left ( \texttt{FRM}(f^{enc}) \right ) \right ]
    \label{eq:4}
\end{equation}
where $\texttt{Re}$ denotes the reshape operation, which is performed using bilinear interpolation. The output of the $n^{th}$ decoder block $B_{dec}^{n}$ is computed as described in (Eq.~\ref{eq:5}).
\begin{equation}
B_{dec}^{n}=\texttt{FRM}\left [ \texttt{BiFFM}  \left \{ \texttt{FRM}(B_{dec}^{n-1}),s_{4} \right \}\right ]\copyright \left [ \texttt{\texttt{Re}}\left ( \texttt{FRM}(B_{dec}^{n-1}) \right ) \right ]
    \label{eq:5}
\end{equation}
 Finally, the predicted mask $f_{out}$ of M-BFF-Net is computed by applying a $1\times1$ convolution operation $f^{1\times1}$ followed by the sigmoid operation $\sigma$ on the last decoder block as given in (Eq.~\ref{eq:6}).
 
\begin{equation}
    f_{out} = \sigma \big(f^{1\times1}(B_{dec}^{4})\big)
    \label{eq:6}
\end{equation}

\subsection{Focal Modulation-based Convformer Attention Block (FMCAB)}

The proposed focal modulation-based convformer attention block (FMCAB) aims to improve segmentation performance. Integrates dynamic attention and adaptive context modulation. Dynamic attention allows the model to focus on crucial regions, while the adaptive context refines the feature representation for those regions. This combination enhances the model's adaptability and is expected to improve segmentation performance in various contexts. Let $in$ be the input to the FMCAB block. The first intermediate output $i_{1}$ of the FMCAB is computed as (Eq.~\ref{eq:7}).
\begin{equation}
    i_{1}=\Re \left ( f^{1\times 1}\left ( f^{3\times 3}\left ( \texttt{LN}\left ( F_{in} \right ) \right ) \right ) \right )
    \label{eq:7}
\end{equation}
where $\texttt{LN}$ is layer normalization, $\Re$ is ReLU activation function, $f^{1\times 1}, f^{3\times 3}$ are the standard convolutions of kernel size $1\times 1$, $3\times 3$, respectively. The second intermediate output $i_{2}$ of the FMCAB is computed as (Eq.~\ref{eq:8}).

\begin{equation}
    i_{2} = \sigma\left ( f^{1\times 1}\left ( \Re \left ( \texttt{GAP} \left ( \texttt{Re}\left ( f^{1\times 1}\left ( i_{1} \right ) \right ) \right )\right ) \right ) \right )\times i_{1}
    \label{eq:8}
\end{equation}
where $\sigma$ is the sigmoid operation and $\texttt{GAP}$ is the global average pooling. The final output of the FMCAB is computed as (Eq.~\ref{eq:9}).

\begin{equation}
    F_{out}= \texttt{FM}\left ( i_{2} \right )+\texttt{LN}\left ( f^{3\times 3}\left ( F_{in} \right ) \right )+f^{1\times 1}\left ( \texttt{Ge}\left ( F_{in} \right )  \right )+i_{1}
    \label{eq:9}
\end{equation}
where $\texttt{Ge}$ is the GeLU activation function, the focal modulation is denoted by $\texttt{FM}$ and computed as (Eq.~\ref{eq:10}).

\begin{equation}
\texttt{FM}= \gamma ^{p}\left [ \sigma\left ( f^{1\times 1}\left ( \texttt{Re}\left ( (\texttt{GAP}(i_{2})-\texttt{GMP}(i_{2}))\times \alpha \right ) \right ) \right )\times i_{2} \right ]
    \label{eq:10}
\end{equation}
where $\gamma^{p}$ is the gamma power and $\alpha$ is the modulation factor.

\begin{figure}[!t]
    \centering
    \includegraphics[width=\textwidth]{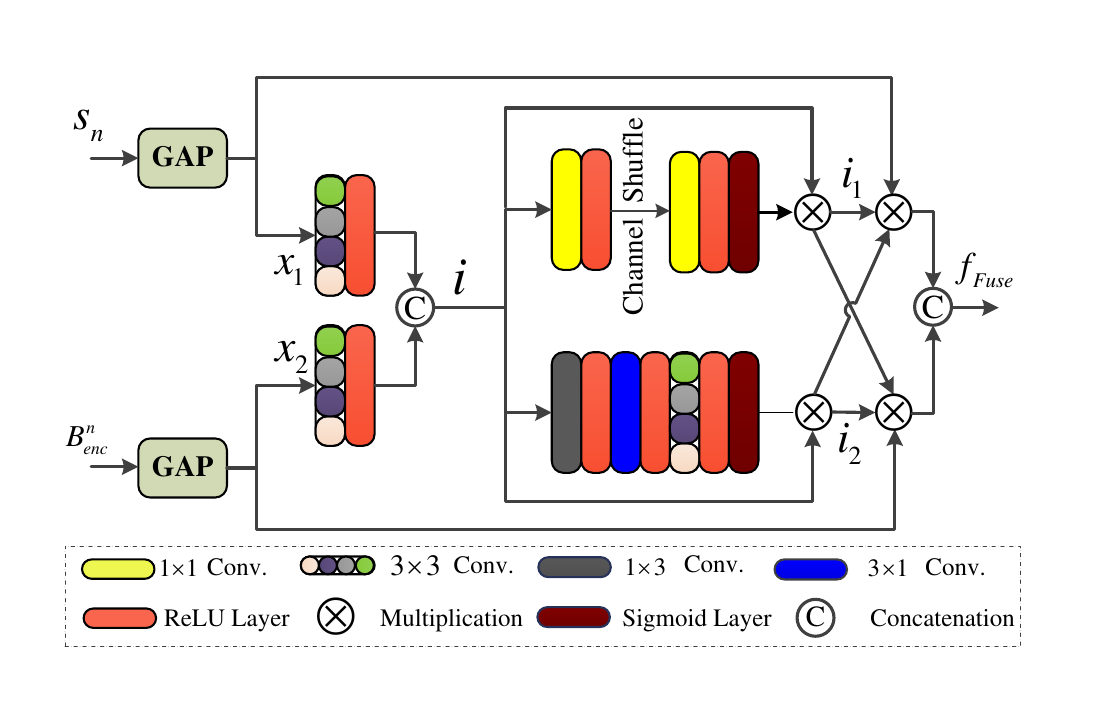}
    \caption{Detailed schematic of the proposed Bidirectional Feature Fusion Module (BiFFM). This module integrates multi-scale features from encoder and decoder branches using global average pooling (GAP) and parallel convolutional pathways. Features $X_1$ and $X_2$
  are first aggregated and then refined through channel-wise operations and channel shuffle to enhance inter-feature interactions. The fusion output $f_{fuse}$ is obtained by concatenating and modulating these feature maps, enabling effective bidirectional information exchange and improved semantic representation.}
    \label{fig:BiFFM}
\end{figure}

\subsection{Bidirectional Feature Fusion Module (BiFFM)}
The proposed M-BFF-Net incorporates a bidirectional feature fusion module (BiFFM), which is pivotal in merging features derived from both skip connections and the decoder-reconstructed feature (Fig.~\ref{fig:BiFFM}). This fusion process harmonizes local and global contexts, enabling the model to achieve a balance between preserving intricate details and recognizing broader contextual information. The BiFFM is a crucial component that contributes significantly to the success of the model by enabling precise and versatile segmentation in diverse medical imaging scenarios.
Let $s_{n}$ and $B_{enc}^{n}$ be the inputs to the BiFFM, and $x_{1}$ and $x_{2}$ the outputs computed by applying global average pooling to these inputs, respectively. The intermediate output $x$ is calculated as in Eq. \ref{Eq:FFM1}).

\begin{equation}
x=\Re \big(f^{1\times1}(x_{1})\big) \copyright \Re \big(f^{1\times1}(x_{2})\big)
 \label{Eq:FFM1}
\end{equation}

which is processed further along two paths. Let $i_{1}$ and $i_{2}$ be the outputs of each path, computed as (Eqs. \ref{Eq:FFM2} - \ref{Eq:FFM3}).
\begin{equation}
{i_{1}} = \sigma \left( {\Re \left( {{f^{1 \times 3}}\left( {\Re \left( {{f^{3 \times 1}}\left( {\Re \left( {{f^{1 \times 1}}\left( x \right)} \right)} \right)} \right)} \right)} \right)} \right){\rm{ }} \times i
 \label{Eq:FFM2}
\end{equation}
and
\begin{equation}
i_{2}=\sigma\left ( \Re\left ( f^{1\times1}\left ( \texttt{CS} \left ( \Re\left ( f^{1\times1}\left ( x \right ) \right ) \right ) \right ) \right ) \right ) \times i
 \label{Eq:FFM3}
\end{equation}

where $\texttt{CS}$ denotes the channel shuffling operation of the convolution layer. The output of the BiFFM is then computed as (Eq. \ref{Eq:FFM4}).
\begin{equation}
f_\text{Fuse} = \big[i_{1} \otimes x_{1} \otimes x_{2}\big] \copyright \big[i_{2} \otimes x_{1} \otimes x_{2}\big].
\label{Eq:FFM4}
\end{equation}

\subsection{Vision Transformer Module (ViTM)}
\label{sec:ViTM}
We incorporate a transformer-based self-attention module, strategically placed in the bottleneck layer. This module capitalizes on the inherent capabilities of transformers to grasp long-range dependencies, facilitate dynamic interactions among features, and improve contextual comprehension. Leveraging transformers and their self-attention mechanisms allows our model to dynamically adjust and enhance feature representations by considering the inherent relationships and dependencies present in medical images. This adaptability is particularly beneficial for addressing complex pathologies and varying lesion sizes, allowing the model to excel in intricate segmentation tasks.

\begin{figure}[!t]
    \centering
    \includegraphics[width=\textwidth]{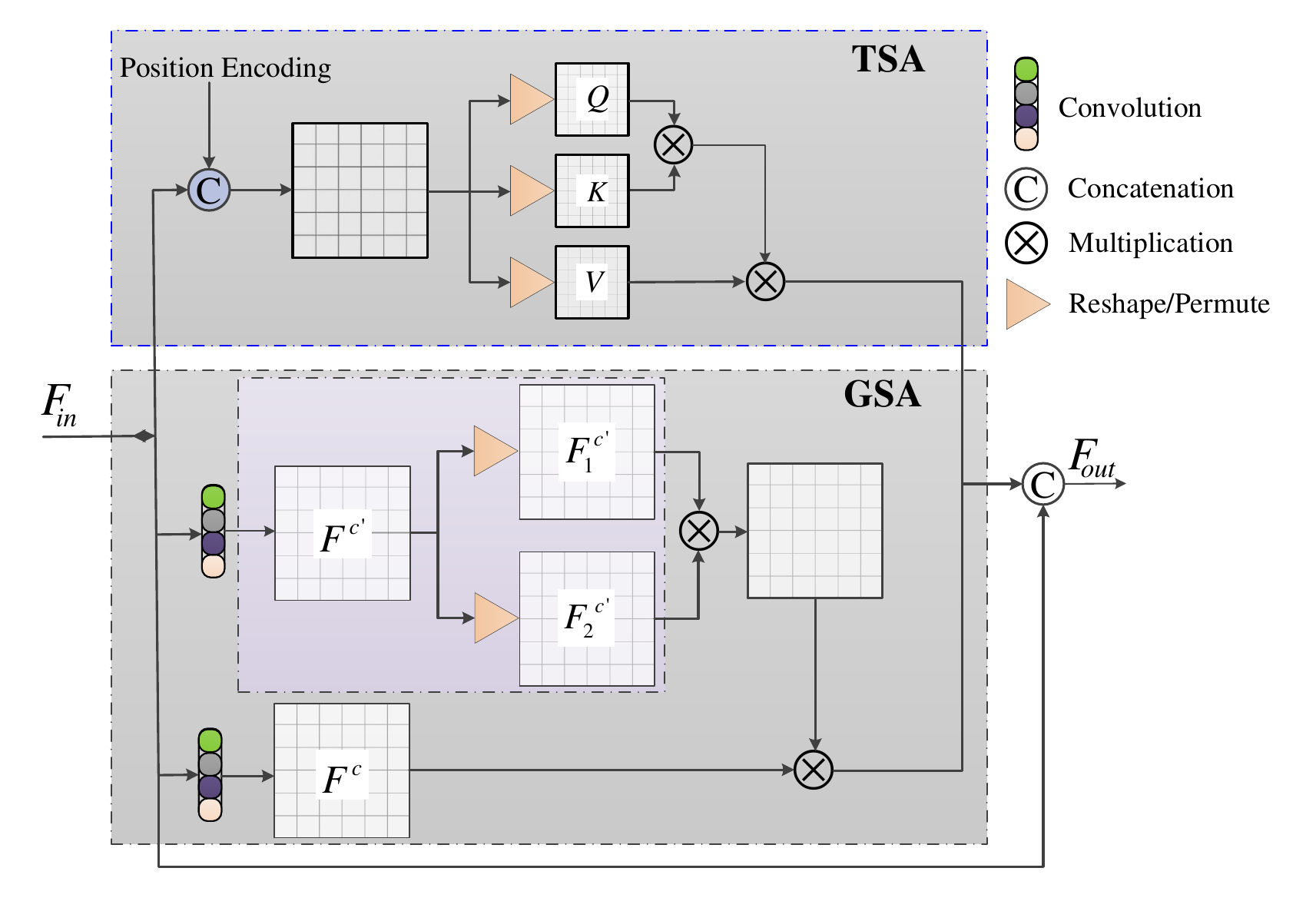}
    \caption{Architectural schematic of the proposed Vision Transformer Module (ViTM). The module consists of two main components: the \textbf{Global Self-Attention (GSA)} mechanism and the \textbf{Token-based Self-Attention (TSA)} module. GSA extracts global contextual relationships through multi-branch attention and hierarchical feature refinement. TSA operates on tokenized representations with positional encoding, utilizing multi-head self-attention via query ($Q$), key ($K$), and value ($V$) embeddings. The final refined feature map $F_{\text{out}}$ is generated through concatenation and permutation operations, allowing for rich global feature learning in medical image segmentation.}
    \label{fig:ViTM}
\end{figure}

ViTM (Fig.~\ref{fig:ViTM}) uses a combination of transformer self-attention (TSA) and global spatial attention (GSA) modules.
The input feature map $F_\text{in}$ is then embedded in three matrices $Q\in\mathbb{R}^{(h\times w)\times c}, K\in\mathbb{R}^{c\times (h\times w)}, V\in\mathbb{R}^{c\times (h\times w)}$, given by
\begin{equation}
    Q =  W_Q\cdot F_\text{in}
\end{equation}
\begin{equation}
    K =  W_K\cdot F_\text{in}
\end{equation}
\begin{equation}
    V =  W_V\cdot F_\text{in}
\end{equation}
where $W_Q, W_K, W_V$ are three embedding functions for different linear projections. The operation of the scaled dot product with Softmax normalization between $Q$ and $K$ gives $S\in\mathbb{R}^{c\times c}$, which represents the similarity between channels in $Q$ and others. To obtain the aggregation values weighted by attention weights, $S$ is multiplied by the value matrix $V$ so that the multihead attention mechanism can be written as 
\begin{equation}
    A_\text{TSA}(Q, K, V) = \text{Softmax}\left(\frac{QK}{\sqrt{d_k}}V\right).
\end{equation}
Finally, $A_\text{TSA}\in\mathbb{R}^{c\times (h\times w)}$ is reshaped to $\mathbb{R}^{h\times w \times c}$, equal to the input shape.

GSA is employed to capture information on global position dependencies. The input feature map $F_\text{in}\in\mathbb{R}^{h\times w\times c}$ is first embedded in $F^c\in\mathbb{R}^{h\times w\times c}$ and $F^{c'}\in\mathbb{R}^{h\times w\times c'}$ where $c' = c/2$. After reshaping $F^{c'}\in\mathbb{R}^{h\times w\times c'}$ to $F^{c'}_1\in\mathbb{R}^{(h\times w)\times c'}$ and $F^{c'}_2\in\mathbb{R}^{c'\times (h\times w)}$, respectively, the scaled dot product of $F^{c'}_1$ and $F^{c'}_2$ then passes to a Softmax normalization layer, where the output map $S\in\mathbb{R}^{(h\times w)\times (h\times w)}$ indicates spatial similarity and $S_{i,j}$ represents the correlation between position $i^\text{th}$ and $j^\text{th}$. The multihead attention mechanism can be written as
\begin{equation}
    A_\text{GSA}(Q,K,V) = \text{Softmax}(F^{c'}_1\cdot F^{c'}_2)F^c 
            = \frac{f^{c'}_1\cdot f^{c'}_2}{F^c}.
\end{equation}

\section{Experiments and Results}
\label{sec:Results} 


\subsection{Datasets}

We conducted experiments using three widely used datasets for polyp segmentation (Kvasir-SEG, CVC-ClinicDB and CVC-ColonDB), for skin lesion segmentation (ISIC2016, ISIC2017 and ISIC2018), and ultrasound image segmentation (BUSI for breast lesion segmentation and DDTI for thyroid nodule segmentation). The datasets are summarized as follows:
\begin{enumerate}
    \item \textbf{Kvasir-SEG:} This dataset consists of 1,000 polyp images along with their corresponding ground truth masks. The image resolutions vary between $332\times 487$ and $1920\times 1072$ pixels.
    \item \textbf{CVC-ClinicDB:} This dataset includes 612 polyp images with their ground truth masks, all at a fixed resolution of $384\times 288$ pixels.
    \item \textbf{CVC-ColonDB:} Contains 380 polyp images, each accompanied by ground truth masks, with images at a resolution of $574\times 500$ pixels.    

    \item \textbf{ISIC 2016:} This dataset contains 900 dermoscopic images in the training set and 379 images in the test set along with their ground-truth masks. The image resolutions vary between $679\times 566$ and $2848\times 4288$ pixels.
    \item \textbf{ISIC 2017:} This dataset offers a larger collection with 2,000 dermoscopic images for training, all of which are paired with the corresponding ground truth masks. Additionally, it includes 150 images for validation and another set of 600 images for assessing the framework's performance. The image resolutions vary between $679\times 453$ and $6748\times 4499$ pixels.
    \item \textbf{ISIC 2018:} The ISIC 2018 dataset comprises 2,594 dermoscopic images designated for training, each with its corresponding ground truth mask. The dataset also provides an additional set of 1,000 images reserved for evaluating the performance of the developed framework. The image resolutions vary between $679\times 453$ to $6748\times 4499$ pixels.
    
    \item \textbf{BUSI:} The BUSI dataset comprises 780 breast ultrasound images collected from women aged between 25 and 75 years of age. The images are in $.png$ format and have an average size of $500\times500$ pixels. Ground truth masks are available for all images.

    \item \textbf{DDTI:} The DDTI dataset consists of 637 ultrasound thyroid nodule images with varying resolutions such as $560\times 360$, $280\times 360$, and $245\times360$. Ground truth masks are available for all images.


\end{enumerate}

\subsection{Experimental Setup and Training Details}
\label{experimentalResults}

In our experiment for polyp segmentation, we used a combined dataset, merging the ClinicDB dataset and the Kvasir-SEG dataset, as outlined in the experimental setup of Meta-Polyp \cite{meta_polyp}. This merged training set is widely adopted in various subsequent methods. It consists of two subsets: Kvasir-SEG (900 training images) and CVC-ClinicDB (550 training images). For benchmarking, we selected three datasets: Kvasir-SEG, ColonDB, and CVC300 dataset. Among these, only Kvasir-SEG is within the distribution, while the remaining two datasets are considered out-of-distribution.  
 
For the BUSI and DDTI datasets, the data were split into training and validation sets in a ratio of $80\%:20\%$. To augment the dataset, we applied rotations ranging from $0^{\circ}$ to $360^{\circ}$ with a step size of $30^{\circ}$, along with brightness adjustments by factors of 0.8 and 1.2. Performance evaluation was performed using 5-fold cross-validation. In the case of segmentation of the skin lesion, the model was trained without data augmentation. We employed a $80\%:20\%$ split for training and validation, while performance was evaluated using the test sets of all three datasets. 


During model training, Adam Optimizer was utilized with a maximum of 100 iterations and an initial learning rate set at 0.001. If the validation set showed no improvement after seven epochs, the learning rate was reduced by 25\%. An early stopping mechanism (after 10 epochs) was applied to mitigate overfitting. The models were developed using Keras, with TensorFlow serving as the back-end, and training was conducted on a NVIDIA K80 GPU.

\subsection{Evaluation Criteria}

Performance quantification was performed using five evaluation metrics: accuracy, sensitivity, specificity, Jaccard index, and Dice coefficient. 
\begin{equation}
\mathrm{Accuracy\ (A_{cc}) = \frac{T_P+T_N}{T_P+T_N+F_P+F_N}},
\end{equation}
\begin{equation}
\mathrm{Sensitivity\ (S_n) = \frac{T_P}{T_P+F_N}},
\end{equation}
\begin{equation}
\mathrm{Specificity\ (S_p) = \frac{T_N}{T_N+F_P}},
\end{equation}
\begin{equation}
\mathrm{Jaccard\ (J) = \frac{T_P}{T_P+F_P+F_N}},
\end{equation}
\begin{equation}
\mathrm{Dice\ (D) = \frac{2T_P}{2T_P+F_P+F_N}}.
\end{equation}
All metrics range from 0 (worst performance) to 1 (best performance).


\subsection{Comparison with SOTA Networks}

\subsubsection{Polyp Segmentation}
The performance of M-BFF-Net for polyp segmentation was evaluated on three publicly available datasets. For polyp segmentation on the Kvasir-SEG dataset, M-BFF-Net was compared with several methods, including
ARU-GD \cite{maji2022attention}, BCDU-Net \cite{azad2019bi}, Duck-Net \cite{duck_net}, Meta-Polyp \cite{meta_polyp}, Swin-Unet \cite{cao2023swin}, TBconvL-Net \cite{iqbal2024tbconvl}, U-Net \cite{ronneberger2015u}, and UNet++ \cite{zhou2018unet++}. As shown in Tables (\ref{tab:Kvasir}--\ref{tab:colon}), the Jaccard index of M-BFF-Net outperformed these methods, achieving improvements from 0.86\% to 18.92\%, on the Kvasir-SEG dataset, 1.68\% to 34.54\% on the CVC-300 dataset, and 0.29\% to 21.27\% on the CVC-ColonDB dataset, respectively.

\begin{table}[!t]
  \centering
  \caption{Performance comparison of M-BFF-Net model with various SOTA methods on Kvasir-SEG dataset.}
  \adjustbox{max width=\textwidth}{%
    \begin{tabular}{lccccc}
    \toprule
    \multirow{2}[4]{*}{\textbf{Method}} & \multicolumn{5}{c}{\textbf{Performance Measures in (\%)}} \\
\cmidrule{2-6}          & $J$ & $D$  & $A_{cc}$   & $S_{n}$   & $P{r}$ \\
    \midrule
    ARU-GD \cite{maji2022attention} & 75.84 & 86.26 & 95.83 & 80.05 & 93.51 \\
    BCDU-Net \cite{azad2019bi} & 74.04 & 85.08 & 95.43 & 80.74 & 89.93 \\
    Duck-Net \cite{duck_net} & 90.51 & 95.02 & 98.42 & 93.79 & 96.28 \\
    Meta-Polyp \cite{meta_polyp} & 92.10 & 95.90 & 97.89 & 93.37 & 93.50 \\
    Swin-Unet \cite{cao2023swin} & 74.38 & 85.30 & 95.39 & 82.97 & 87.78 \\
    TBconvL-Net \cite{iqbal2024tbconvl} & 85.54 & 92.20 & 97.49 & 92.03 & 92.38 \\
    U-Net \cite{ronneberger2015u}  & 76.29 & 86.55 & 95.63 & 87.18 & 85.93 \\
    UNet++ \cite{zhou2018unet++} & 83.39 & 90.94 & 97.38 & 86.40 & 95.99 \\
    \midrule
    \textbf{M-BFF-Net} & \textbf{92.96} & \textbf{95.14} & \textbf{98.78} & \textbf{94.69} & \textbf{97.38} \\
    \bottomrule
    \end{tabular}%
    }
  \label{tab:Kvasir}%
\end{table}%

\begin{figure*}[h]
    \centering
    \includegraphics[width=1\textwidth]{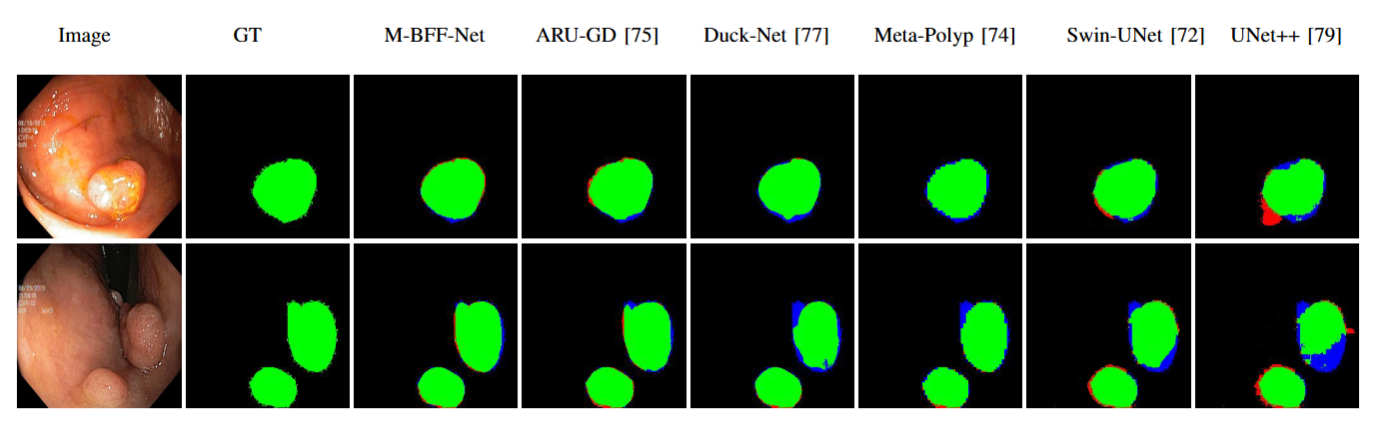}
    \caption{ Visual performance comparison of the proposed M-BFF-Net on Kvasir-SEG dataset.}
    \label{fig:kvasir}
\end{figure*}

\begin{table}[!t]
  \centering
  \caption{Performance comparison of M-BFF-Net model with various SOTA methods on CVC-300 dataset.}
  \adjustbox{max width=\textwidth}{%
    \begin{tabular}{lccccc}
    \toprule
    \multirow{2}[4]{*}{\textbf{Method}} & \multicolumn{5}{c}{\textbf{Performance Measures in (\%)}} \\
\cmidrule{2-6}          & $J$ & $D$  & $A_{cc}$   & $S_{n}$   & $P{r}$ \\
    \midrule
    ARU-GD \cite{maji2022attention} & 77.66 & 87.42 & 97.80 & 80.20 & 96.08 \\
    BCDU-Net \cite{azad2019bi} & 57.23 & 72.79 & 95.86 & 61.37 & 89.45 \\
    Duck-Net \cite{duck_net} & 90.09 & 94.78 & 99.07 & 94.89 & 94.68 \\
    Meta-Polyp \cite{meta_polyp} & 89.52 & 94.50 & 99.03 & 94.06 & 94.88 \\
    Swin-Unet \cite{cao2023swin} & 82.82 & 90.60 & 98.42 & 86.21 & 95.47 \\
    TBconvL-Net \cite{iqbal2024tbconvl} & 87.40 & 93.27 & 98.86 & 89.58 & 97.28 \\
    U-Net \cite{ronneberger2015u}  & 61.69 & 76.31 & 95.99 & 73.03 & 79.89 \\
    UNet++ \cite{zhou2018unet++} & 63.61 & 77.76 & 96.29 & 73.46 & 82.60 \\
    \midrule
    \textbf{M-BFF-Net} & \textbf{91.77} & \textbf{96.04} & \textbf{99.13} & \textbf{94.86} & \textbf{97.73} \\
    \bottomrule
    \end{tabular}%
    }
  \label{tab:CVC}%
\end{table}%

\begin{figure*}[h]
    \centering
    \includegraphics[width=1\textwidth]{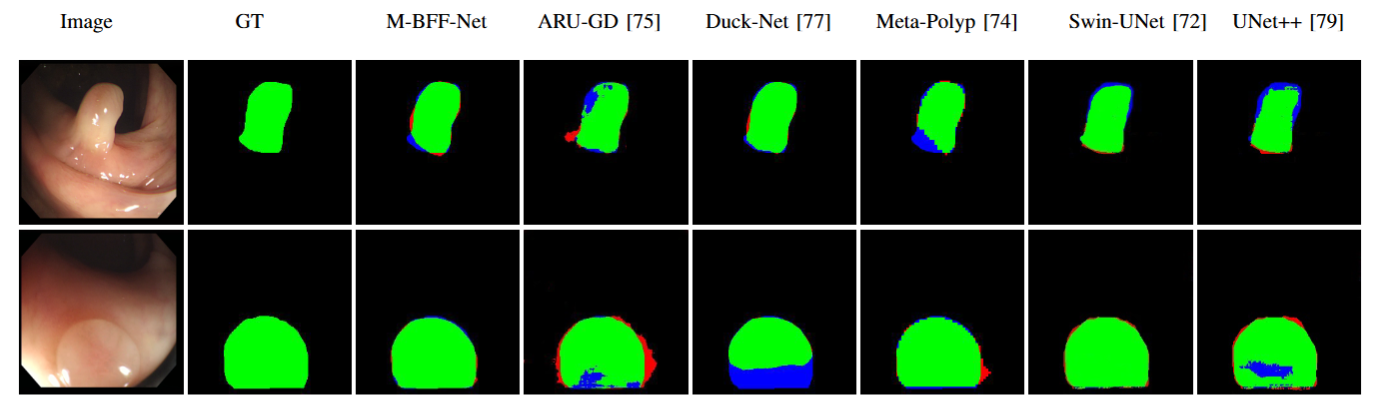}
    \caption{Visual performance comparison of the proposed M-BFF-Net on CVC-Clinic dataset.}
    \label{fig:CVC}
\end{figure*}

\begin{figure*}[h]
    \centering
    \includegraphics[width=1\textwidth]{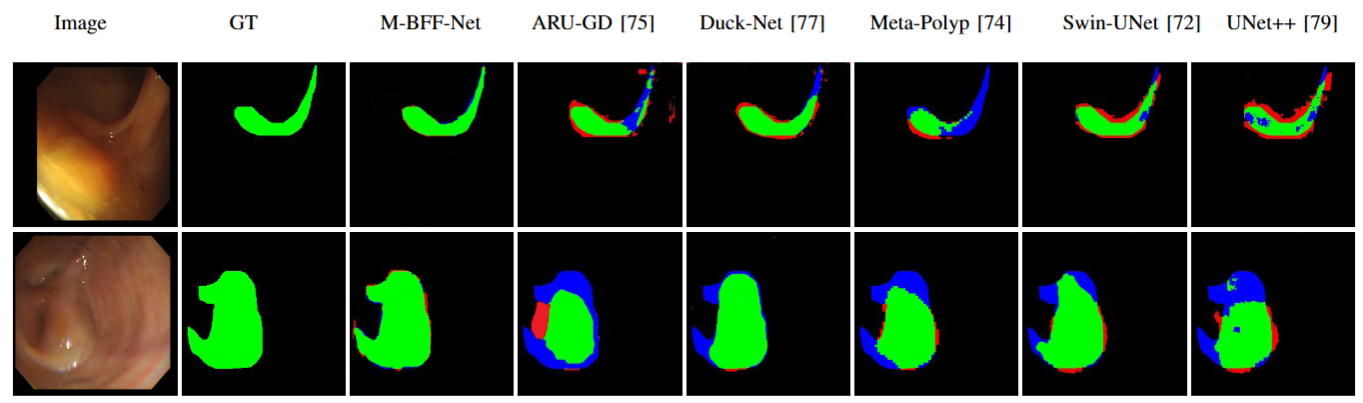}
    \caption{Visual performance comparison of the proposed M-BFF-Net on CVC-ColonDB dataset.}
    \label{fig:colon}
\end{figure*}

Figures (\ref{fig:kvasir}-\ref{fig:colon}) present the visual results of the proposed M-BFF-Net with other methods including ARU-GD \cite{maji2022attention}, Duck-Net \cite{duck_net}, Meta-Polyp \cite{meta_polyp}, Swin-Unet \cite{cao2023swin} and UNet++ \cite{zhou2018unet++}. In all datasets, M-BFF-Net demonstrated superior segmentation capabilities, producing the best segmentation results that closely align with GT data, particularly when dealing with complex cases involving low contrast, multiple lesions, irregular shapes, and size variations. The results indicate that M-BFF-Net is highly effective in accurately segmenting challenging polyp images.

\begin{table}[!t]
  \centering
  \caption{Performance comparison of M-BFF-Net model with various SOTA methods on CVC-ColonDB dataset.}
  \adjustbox{max width=\textwidth}{%
    \begin{tabular}{lccccc}
    \toprule
    \multirow{2}[4]{*}{\textbf{Method}} & \multicolumn{5}{c}{\textbf{Performance Measures in (\%)}} \\
\cmidrule{2-6}          & $J$ & $D$  & $A_{cc}$   & $S_{n}$   & $P{r}$ \\
    \midrule
    ARU-GD \cite{maji2022attention} & 84.01 & 91.31 & 99.01 & 86.59 & \textbf{96.57} \\
    BCDU-Net \cite{azad2019bi} & 68.70 & 73.98 & 97.61 & 60.57 & 95.00 \\
    Duck-Net \cite{duck_net} & 85.71 & 92.30 & 99.14 & 93.51 & 91.13 \\
    Meta-Polyp \cite{meta_polyp} & 87.85 & 93.53 & \textbf{99.29} & 93.92 & 93.14 \\
    Swin-Unet \cite{cao2023swin} & 71.98 & 83.71 & 98.29 & 81.51 & 86.03 \\
    TBconvL-Net \cite{iqbal2024tbconvl} & 83.04 & 90.73 & 98.99 & 90.40 & 91.07 \\
    U-Net \cite{ronneberger2015u}  & 70.37 & 80.32 & 98.07 & 82.74 & 81.00 \\
    UNet++ \cite{zhou2018unet++} & 66.87 & 63.83 & 95.65 & 70.10 & 58.58 \\
    \midrule
    \textbf{M-BFF-Net} & \textbf{88.14} & \textbf{93.67} & 99.20 & \textbf{94.90} & 93.53 \\
    \bottomrule
    \end{tabular}%
    }
  \label{tab:colon}%
\end{table}%

\subsubsection{Skin Lesion Segmentation}

\begin{figure*}[h]
    \centering
    \includegraphics[width=1\textwidth]{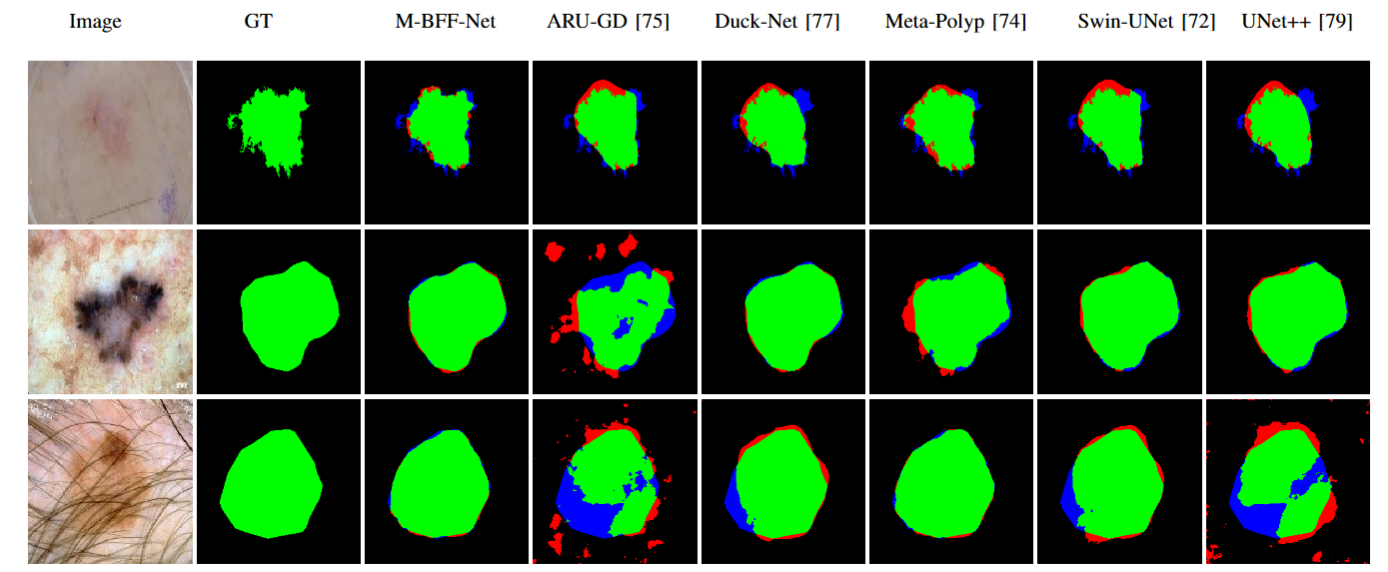}
    \caption{Visual performance comparison of the proposed M-BFF-Net on CVC-ColonDB dataset.}
    \label{fig:skin}
\end{figure*}

M-BFF-Net performance was evaluated on three publicly available datasets for skin lesion segmentation (ISIC2016, ISIC2017, and ISIC 2018). For segmentation of skin lesion, M-BFF-Net was compared with several methods, including ARU-GD \cite{maji2022attention}, BCDU-Net \cite{azad2019bi}, Duck-Net \cite{duck_net}, Meta-Polyp \cite{meta_polyp}, Swin-Unet \cite{cao2023swin}, TBconvL-Net \cite{iqbal2024tbconvl}, U-Net \cite{ronneberger2015u} and UNet++ \cite{zhou2018unet++}. The results (Table \ref{tab:skin}) show that M-BFF-Net consistently outperformed the existing approaches in the ISIC 2016, 2017, and 2018 datasets, achieving Jaccard score improvements of 0. 81\% –8. 9\%, 0. 5\% –9. 61\% and 0. 05\% –11. 61\%, respectively. Visual comparisons (Fig. \ref{fig:skin}) further validate its superiority, particularly in challenging cases involving occlusions, black backgrounds, hair, low contrast, varying size of the lesion and irregular boundaries.

\begin{table*}[htbp]
  \centering
  \caption{Performance comparison of M-BFF-Net model with various SOTA methods on skin lesion segmentation datasets (ISIC2016, ISIC2017, ISIC2018).}
  \adjustbox{max width=\textwidth}{%
    \begin{tabular}{lccccccccccccccccc}
    \toprule
    \multirow{3}[4]{*}{\textbf{Method}} & \multicolumn{17}{c}{\textbf{Performance Measures in (\%)}} \\
\cmidrule{2-18}                    & \multicolumn{5}{c}{\textbf{ISIC2018}} &       & \multicolumn{5}{c}{\textbf{ISIC2017}} &       & \multicolumn{5}{c}{\textbf{ISIC2016}} \\
\cmidrule{2-6}\cmidrule{8-12}\cmidrule{14-18}        & $J$ & $D$  & $A_{cc}$   & $S_{n}$   & $S{p}$ &       & $J$ & $D$  & $A_{cc}$   & $S_{n}$   & $S{p}$ &       & $J$ & $D$  & $A_{cc}$   & $S_{n}$   & $S{p}$ \\
    \midrule
    ARU-GD \cite{maji2022attention} & 84.55 & 89.16 & 94.23 & 91.42 & 96.81 &       & 80.77 & 87.89 & 93.88 & 88.31 & 96.31 &       & 85.12 & 90.83 & 94.38 & 89.86 & 94.65 \\
    BCDU-Net \cite{azad2019bi} & 81.10 & 85.10 & 93.70 & 78.50 & 98.20 &       & 79.20 & 78.11 & 91.63 & 76.46 & 97.09 &       & 83.43 & 80.95 & 91.78 & 78.11 & 96.20 \\
    Duck-Net \cite{duck_net} & 81.13 & 88.07 & 93.24 & 90.72 & 95.88 &       & 75.94 & 84.25 & 93.26 & 83.63 & 97.25 &       & 84.27 & 89.95 & 95.67 & 93.14 & 94.68 \\
    FAT-Net \cite{WU2022102327} & 82.02 & 89.03 & 95.78 & 91.00 & 96.99 &       & 76.53 & 85.00 & 93.26 & 83.92 & 97.25 &       & 85.30 & 91.59 & 96.04 & 92.59 & 96.02 \\
    Meta-Polyp \cite{meta_polyp} & 83.76 & 90.41 & 97.24 & 91.66 & 98.63 &       & 79.88 & 87.69 & 94.96 & 89.53 & 96.55 &       & 83.81 & 90.23 & 95.09 & 92.11 & 95.91 \\
    RA-Net \cite{HU2022117112}  & 83.09 & 89.55 & 95.68 & 93.06 & 94.69 &       & 80.51 & 88.07 & 94.66 & 89.92 & 95.72 &       & 84.27 & 89.95 & 95.67 & 93.14 & 94.68 \\
    Swin-Unet \cite{cao2023swin} & 82.79 & 88.98 & 96.83 & 90.10 & 97.16 &       & 80.89 & 81.99 & 94.76 & 88.06 & 96.05 &       & 87.60 & 88.94 & 96.00 & 92.27 & 95.79 \\
    TBconvL-Net \cite{iqbal2024tbconvl} & 91.65 & \textbf{95.47} & 97.60 & 95.29 & 98.55 &       & 84.80 & 90.89 & 96.07 & 91.19 & 97.61 &       & 89.47 & 95.45 & 97.05 & 94.02 & 97.68 \\
    U-Net \cite{ronneberger2015u}  & 80.09 & 86.64 & 92.52 & 85.22 & 92.09 &       & 75.69 & 84.12 & 93.29 & 84.30 & 93.41 &       & 81.38 & 88.24 & 93.31 & 87.28 & 92.88 \\
    UNet++ \cite{zhou2018unet++} & 81.62 & 87.32 & 93.72 & 88.70 & 93.96 &       & 78.58 & 86.35 & 93.73 & 87.13 & 94.41 &       & 82.81 & 89.19 & 93.88 & 88.78 & 93.52 \\
    \midrule
    \textbf{M-BFF-Net} & \textbf{91.70} & 94.47 & \textbf{97.84} & \textbf{95.84} & \textbf{98.52} &       & \textbf{85.30} & \textbf{91.15} & \textbf{96.34} & \textbf{91.77} & \textbf{97.73} &       & \textbf{90.28} & \textbf{96.28} & \textbf{97.39} & \textbf{94.85} & \textbf{98.46} \\
    \bottomrule
    \end{tabular}%
    }
  \label{tab:skin}%
\end{table*}%

\subsubsection{Ultrasound Image Segmentation}

The performance of M-BFF-Net was evaluated on two publicly available datasets: the BUSI dataset for breast cancer segmentation and the DDTI dataset for thyroid nodule segmentation. For breast cancer segmentation, M-BFF-Net was compared with several methods, including ARU-GD \cite{maji2022attention}, BCDU-Net \cite{azad2019bi}, Duck-Net \cite{duck_net}, Meta-Polyp \cite{meta_polyp}, Swin-Unet \cite{cao2023swin}, TBconvL-Net \cite{iqbal2024tbconvl}, U-Net \cite{ronneberger2015u} and UNet++ \cite{zhou2018unet++}. As shown in Table \textbf{\ref{tab:U1}}, the Jaccard index of M-BFF-Net outperformed these methods, achieving improvements from 0. 64\% to 11. 35\% on the BUSI dataset.

\begin{table}[!t]
  \centering
  \caption{Performance (mean ± std) comparison of M-BFF-Net model with various SOTA methods on the breast lesion segmentation dataset BUSI \cite{BUSIdataset}.}
    \adjustbox{max width=\textwidth}{%
    \begin{tabular}{lccccc}
    \toprule
    \multirow{2}[4]{*}{\textbf{Method}} & \multicolumn{5}{c}{\textbf{Performance (\%)}} \\
    \cmidrule{2-6} & $J$ & $D$  & $A_{cc}$   & $S_{n}$   & $S{p}$ \\
    \midrule
    ARU-GD \cite{maji2022attention} & 77.07$\pm$2.96 & 83.64$\pm$2.53 & 97.94$\pm$1.32 & 83.80$\pm$1.87 & 98.78$\pm$2.59 \\
    BCDU-Net \cite{azad2019bi} & 74.49$\pm$3.65 & 66.75$\pm$2.31 & 94.82$\pm$1.28 & 86.85$\pm$3.95 & 95.57$\pm$2.72 \\
    Duck-Net \cite{duck_net} & 77.48$\pm$2.08 & 84.68$\pm$2.91 & 97.82$\pm$1.92   & 85.37$\pm$3.32 & 98.14$\pm$1.95 \\
    Meta-Polyp \cite{meta_polyp} & 75.97$\pm$3.81 & 83.97$\pm$2.20 & 96.22$\pm$2.85   & 83.45$\pm$2.28 & 97.11$\pm$3.80\\
    Swin-UNet \cite{cao2023swin} & 77.16$\pm$2.85 & 84.45$\pm$2.54 & 97.55$\pm$2.77 & 84.81$\pm$3.99 & 98.34$\pm$2.50 \\   
    TBconvL-Net \cite{iqbal2024tbconvl} & 76.09$\pm$2.04 & 83.67$\pm$3.06 & 96.65$\pm$2.55     & 84.39$\pm$1.13 & 98.04$\pm$1.44\\     
    U-Net \cite{ronneberger2015u} & 67.77$\pm$2.35 & 76.96$\pm$3.67 & 95.48$\pm$3.60 & 78.33$\pm$4.35 & 96.13$\pm$2.07 \\
    UNet++ \cite{zhou2018unet++} & 76.85$\pm$3.13 & 76.22$\pm$3.59 & 97.97$\pm$1.93 & 78.61$\pm$3.27 & 98.86$\pm$2.23 \\
  
    \midrule
    \textbf{Proposed M-BFF-Net} & \textbf{79.12$\pm$1.85} & \textbf{85.42$\pm$2.13} & \textbf{98.04$\pm$1.10} & \textbf{87.93$\pm$1.02} & \textbf{98.72$\pm$1.26} \\
    \bottomrule
    \end{tabular}%
  }
  \label{tab:U1}
\end{table}

\begin{figure*}[h]
    \centering
    \includegraphics[width=1\textwidth]{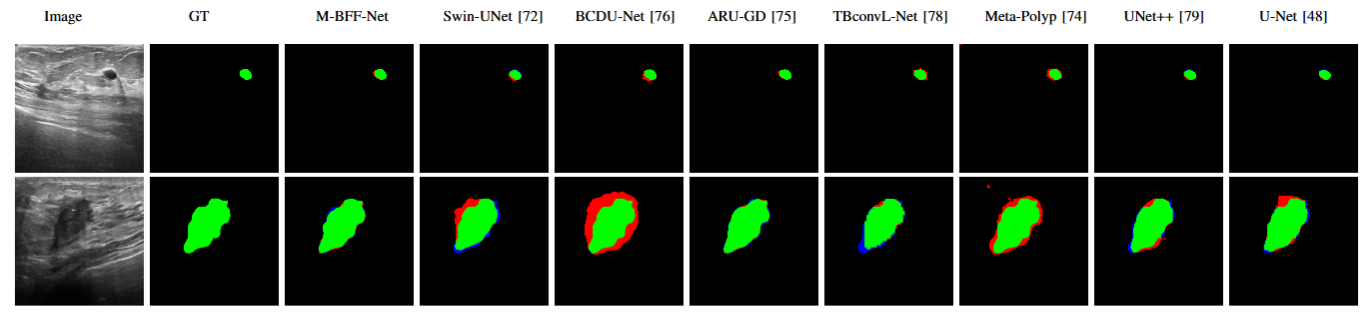}
    \caption{Visual performance comparison of the proposed M-BFF-Net on BUSI \cite{BUSIdataset} dataset.}
    \label{fig:U1}
\end{figure*}

Similarly, for thyroid nodule segmentation on the DDTI dataset, M-BFF-Net was evaluated against multiple SOTA models, including ARU-GD \cite{maji2022attention}, BCDU-Net \cite{azad2019bi}, Duck-Net \cite{duck_net}, Meta-Polyp \cite{meta_polyp}, Swin-Unet \cite{cao2023swin}, TBconvL-Net \cite{iqbal2024tbconvl}, U-Net \cite{ronneberger2015u}, and UNet++ \cite{zhou2018unet++}. As reported in Table \ref{tab:U2}, the Jaccard index of M-BFF-Net showed notable improvements from 1.3\% to 11.67\% compared to these methods on the DDTI dataset.

\begin{table}[!t]
    \centering
    \caption{Performance (mean ± std) comparison of M-BFF-Net with various SOTA methods on the thyroid nodule segmentation dataset DDTI \cite{DDTI}.}
    \adjustbox{max width=\textwidth}{%
    \begin{tabular}{lccccc}
    \toprule
    \multirow{2}[4]{*}{\textbf{Method}} & \multicolumn{5}{c}{\textbf{Performance (\%)}} \\
    \cmidrule{2-6}          & $J$ & $D$  & $A_{cc}$   & $S_{n}$   & $S_{p}$ \\
    \midrule
    ARU-GD \cite{maji2022attention} & 77.07$\pm$1.90 & 83.64$\pm$2.56 & 97.94$\pm$2.55 & 83.80$\pm$4.20 & 98.78$\pm$2.35 \\
    BCDU-Net \cite{azad2019bi} & 77.79$\pm$1.90 & 79.49$\pm$2.27 & 93.22$\pm$2.51 & 82.31$\pm$3.39 & 94.34$\pm$1.34 \\
    Duck-Net \cite{duck_net} & 83.43$\pm$1.87 & 86.01$\pm$1.78 & 97.98$\pm$2.30 & 82.21$\pm$3.07 & 98.88$\pm$1.13  \\   
   Meta-Polyp \cite{meta_polyp} & 80.76$\pm$2.42 & 85.59$\pm$1.80 & 97.79$\pm$2.65 & 85.23$\pm$3.74 & 98.98$\pm$2.01  \\
    Swin U-Net \cite{cao2023swin} & 83.44$\pm$2.49 & 86.86$\pm$2.45 & 96.93$\pm$2.18 & 86.42$\pm$2.39 & 97.98$\pm$2.05\\
    TBconvL-Net \cite{iqbal2024tbconvl} & 82.66$\pm$2.14 & 85.72$\pm$1.02 & 97.91$\pm$2.60 & 79.54$\pm$4.28 & 98.82$\pm$2.18  \\
    U-Net \cite{ronneberger2015u}  & 74.76$\pm$1.36 & 84.08$\pm$3.19 & 96.55$\pm$2.48 & 85.50$\pm$3.09 & 97.57$\pm$1.61 \\
    UNet++ \cite{zhou2018unet++} & 74.76$\pm$3.46 & 84.08$\pm$2.27 & 96.55$\pm$2.51 & 85.50$\pm$3.39 & 97.57$\pm$1.34 \\

    \midrule
    \textbf{Proposed M-BFF-Net} & \textbf{85.73$\pm$1.19} & \textbf{89.01$\pm$1.01} & \textbf{98.15$\pm$1.24} & \textbf{88.13$\pm$1.18} & \textbf{99.05$\pm$1.03} \\
    \bottomrule
    \end{tabular}%
    }
    \label{tab:U2}
\end{table}

\begin{figure*}[h]
    \centering
    \includegraphics[width=1\textwidth]{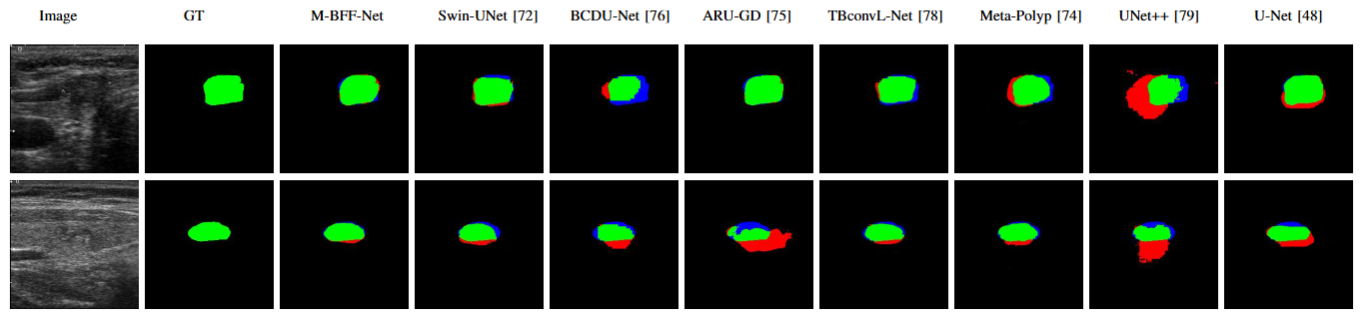}
    \caption{Visual performance comparison of the proposed M-BFF-Net on DDTI \cite{DDTI} dataset.}
    \label{fig:U2}
\end{figure*}

Figures (\ref{fig:U1}-\ref{fig:U2}) present the visual results of the proposed M-BFF-Net with other methods including ARU-GD \cite{maji2022attention}, Duck-Net \cite{duck_net}, Meta-Polyp \cite{meta_polyp}, Swin-Unet \cite{cao2023swin}, TBconvL-Net \cite{iqbal2024tbconvl}, and UNet++ \cite{zhou2018unet++}. In both datasets, M-BFF-Net demonstrated superior segmentation capabilities, producing results that closely align with the GT data, particularly when dealing with complex cases involving irregular shapes and size variations. The results indicate that M-BFF-Net is highly effective in accurately segmenting challenging ultrasound images.

\subsubsection{Limitations of the Proposed M-BFF-Net}

\begin{figure*}[h]
    \centering
    \includegraphics[width=1\textwidth]{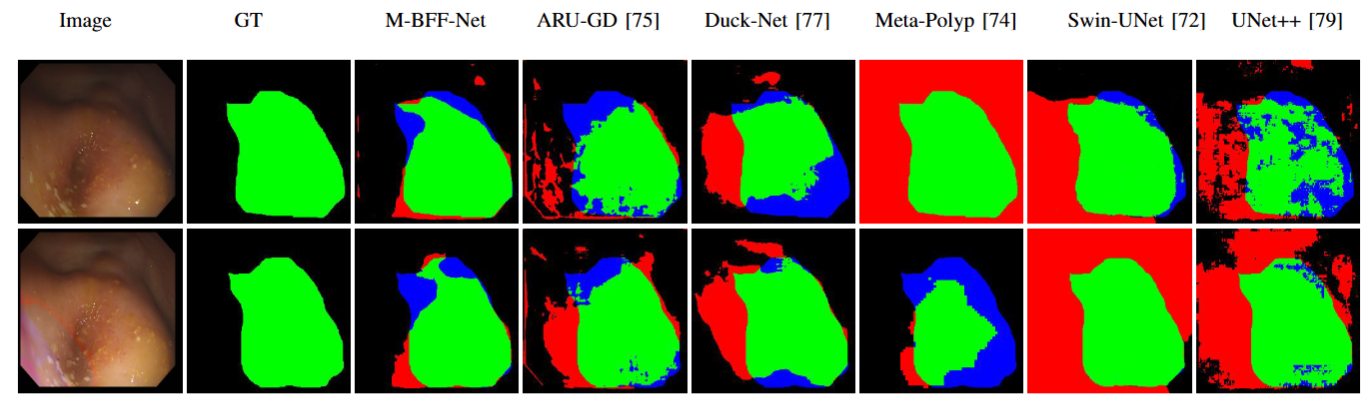}
    \caption{Failure cases comparison of the proposed M-BFF-Net on CVC-ColonDB dataset.}
    \label{fig:colon_fail}
\end{figure*}

\begin{figure*}[h]
    \centering
    \includegraphics[width=1\textwidth]{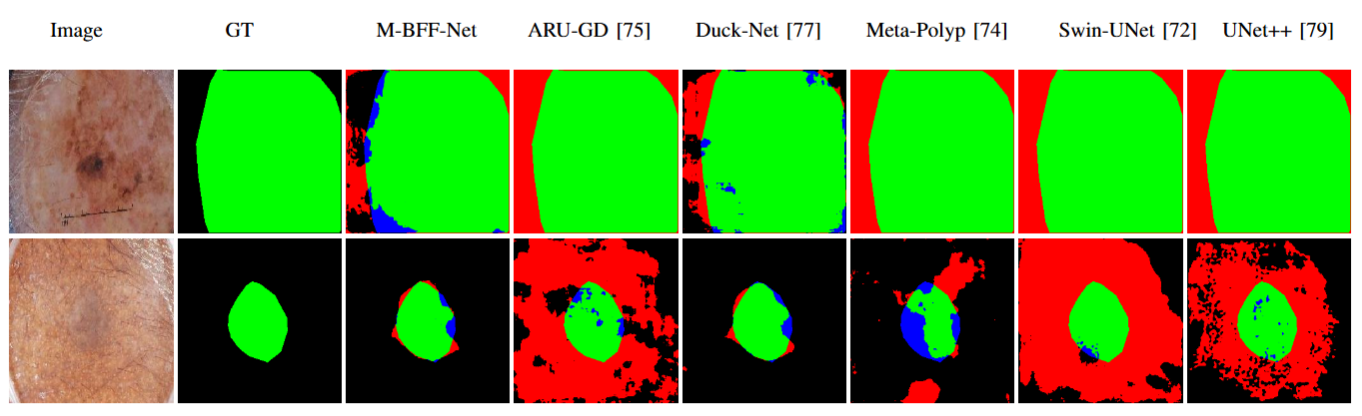}
    \caption{Failure cases comparison of the proposed M-BFF-Net on ISIC2017 dataset.}
    \label{fig:skin_fail}
\end{figure*}

The proposed M-BFF-Net generally performs better compared to existing state-of-the-art techniques; however, it faces limitations in certain situations. These limitations are particularly noticeable in images where there is low contrast between the lesions and the surrounding healthy tissue. As shown in Figures (\ref{fig:colon_fail}-\ref{fig:skin_fail}), accurately defining the boundaries of the lesion becomes more difficult for M-BFF-Net and other methods under these conditions. However, M-BFF-Net exhibits higher segmentation efficiency than its counterparts, marking it as a significant improvement in polyp segmentation, with enhanced outcomes even in challenging cases.

\section{Conclusions}
\label{sec:Conclusions}
In this paper, we presented M-BFF-Net, a novel hybrid deep learning framework that integrates the strengths of both transformer-based attention mechanisms and convolutional neural networks (CNN) for medical image segmentation. M-BFF-Net addresses key limitations in conventional CNN-based models by incorporating two major components: the Focal Modulation-Based ConvFormer Attention Block (FMCAB) and the Bidirectional Feature Fusion Module (BiFFM). This architecture enables the model to effectively capture fine-grained local spatial details as well as global contextual dependencies, which are crucial for accurate medical image segmentation.

The effectiveness of M-BFF-Net was validated in multiple challenging medical imaging tasks, including polyp, skin lesion, and ultrasound image segmentation. Extensive experiments were conducted on a variety of publicly available datasets (e.g. Kvasir-SEG, ISIC2016-2018, BUSI) and M-BFF-Net consistently outperformed existing state-of-the-art (SOTA) methods in key evaluation metrics such as Jaccard index, Dice coefficient, accuracy, sensitivity and specificity. In particular, M-BFF-Net achieved performance gains ranging from 0. 05\% to more than 34\%, demonstrating its robustness in segmenting complex anatomical structures with significant variability in shape, size, and texture.

Although M-BFF-Net demonstrated strong performance even in complex cases with occlusions, irregular lesion boundaries, and low-contrast regions, some limitations were observed in accurately segmenting lesions with extremely subtle boundaries. Moving forward, future work could explore adapting M-BFF-Net to 3D volumetric and multimodal medical images to enhance its spatial understanding and applicability in complex diagnostic scenarios. Additionally, integrating uncertainty quantification mechanisms would allow the model to provide confidence estimates alongside predictions, contributing to more reliable and clinically interpretable segmentation outcomes. In general, M-BFF-Net offers a robust and adaptable solution for medical image segmentation, with the potential for a significant integration into clinical workflows.


\end{document}